\documentclass[runningheads]{llncs}
\usepackage{graphicx}
\usepackage[table]{xcolor}

\usepackage{cite}
\begin{document}
\title{Neural Architecture Search using Particle Swarm and Ant Colony Optimization}
%
%

\author{Séamus Lankford\inst{1} \and Diarmuid Grimes\inst{2}}

%
\institute{Adapt Centre, Dublin City University, Ireland\\ 
\email{seamus.lankford@adaptcentre.ie} \and
Cork Institute of Technology, Ireland\\
\email{diarmuid.grimes@cit.ie}
}

\authorrunning{S. Lankford and D. Grimes}
\titlerunning{OpenNAS using SI}
%

%
\maketitle              
\begin{abstract}
\par
Neural network models have a number of hyperparameters that must be chosen along with their architecture. This can be a heavy burden on a novice user, choosing which architecture and what values to assign to parameters. In most cases, default hyperparameters and architectures are used. Significant improvements to model accuracy can be achieved through the evaluation of multiple architectures. A process known as Neural Architecture Search (NAS) may be applied to automatically evaluate a large number of such architectures.
\par
A system integrating open source tools for Neural Architecture Search (OpenNAS), in the classification of images, has been developed as part of this research. OpenNAS takes any dataset of grayscale, or RBG images, and generates Convolutional Neural Network (CNN) architectures based on a range of metaheuristics using either an AutoKeras, a transfer learning or a Swarm Intelligence (SI) approach.  
\par
Particle Swarm Optimization (PSO) and Ant Colony Optimization (ACO) are used as the SI algorithms. Furthermore, models developed through such metaheuristics may be combined using stacking ensembles.
\par
In the context of this paper, we focus on training and optimizing CNNs using the Swarm Intelligence (SI) components of OpenNAS. Two major types of SI algorithms, namely PSO and ACO, are compared to see which is more effective in generating higher model accuracies. It is shown, with our experimental design, that the PSO algorithm performs better than ACO. The performance improvement of PSO is most notable with a more complex dataset. As a baseline, the performance of fine-tuned pre-trained models is also evaluated.

  
\end{abstract}
\section{Introduction}
The area of Auto Machine Learning (AutoML) \cite{hutter2019automated} is a growing area of interest in recent years. This is reflected in the development of several open source AutoML libraries among which include Auto-WEKA \cite{kotthoff2017auto}, Hyperopt-Sklearn \cite{komer2019hyperopt}, AutoKeras \cite{jin2019auto}, Auto-Sklearn \cite{feurer2019auto, feurer2020auto} and TPOT \cite{olson2019tpot}.   

\par
Despite a renewal of interest in AutoML, many of these open source solutions focus on creating simpler neural architectures. Libraries which concentrate on generating more complex architectures, such as CNNs, are at early stages of development. Consequently they are poorly documented and often unreliable \cite{jin2019auto}. In addition, the alternative of using commercial platforms is expensive and therefore users are left with few practical or viable options.  
\par
The development of OpenNAS integrates several metaheuristic approaches in a single application used for the neural architecture search of more complex neural architectures such as convolutional neural networks.  Furthermore, the effectiveness of NAS in generating good neural architectures for image classification is evaluated.  Standard approaches to NAS, using the AutoKeras framework, are also incorporated into the system design.
\par
A key aspect of the study is to contrast Swarm Intelligence (SI) algorithms for NAS. Consequently,  Particle Swarm Optimization (PSO) \cite{garro2015designing} and Ant Colony Optimization (ACO) \cite{mavrovouniotis2015training} have been chosen as metaheuristics for creating high performing CNN architectures for grayscale and RGB image datasets.

\section{Background}

\subsection{Convolutional Neural Networks}
CNNs are feed-forward Deep Neural Networks (DNNs) used for image recognition. The original CNN architecture was proposed by LeCun \cite{lecun1998gradient} and consisted of two convolution layers, two pooling layers, two fully connected (FC) layers and an output layer. Subsequently, numerous models were developed including popular ones such as ResNet \cite{he2016deep} and VGG \cite{simonyan2014very}. In this study, custom CNN architectures are created by using SI heuristics to find better combinations of convolutional, pooling and FC layers.

\subsection{Auto ML}
AutoML involves the automation of the entire machine learning pipeline including data augmentation, feature engineering, model selection, choice of hyperparameters and finally neural architecture selection and creation. By constrast, NAS has a more narrow focus in that it concentrates on neural architecture selection and creation \cite{elsken2019neural}.   
\par
Tree-based Pipeline Optimization Tool (TPOT)  is an open source python package that uses genetic programming in optimizing the machine learning pipeline \cite{olson2019tpot}.  The library performs well on simple NAS tasks involving the scikit-learn API.  Given this study involves generating more complex CNNs, rather than developing optimal pipelines, it was decided not to use TPOT as part of the initial solution architecture. However, as part of future work, it may have a role in optimizing hyper parameter selection.
\par
AutoKeras \cite{jin2019auto} is an open source AutoML system using Bayesian optimization and network morphism for efficient neural architecture search. 

\subsection{Neural Architecture Search}

\par
Neural architecture search is the process of automatically finding and tuning DNNs. It has been shown that DNNs have made remarkable progress in solving many real world problems such as image recognition, speech recognition and machine translation\cite{sze2017efficient}. In general, NAS systems consist of three main components: a search space, a search algorithm and an evaluation strategy. The search space sets out which architectures can be used in principle whereas the search strategy outlines how the search space is explored. Finally the evaluation strategy determines which architectures yield the best results on unseen data.
\par
A basic approach to NAS is the brute force training and evaluation of all possible model combinations. On completion, the best performing model is selected. However, this is impractical due to the combinatorics of the problem. Using metaheuristics, such as swarm intelligence, is an alternative which seeks the best model within reasonable time constraints.

\subsection{Swarm Intelligence}
 
\par
Swarm Intelligence, a category of Evolutionary Computing, has been used for classification problems in the following forms: Particle Swarm Optimization (PSO) \cite{kennedy1995particle, eberhart1998comparison} and Ant Colony Optimization (ACO) \cite{dorigo1997ant}.

\subsubsection{Particle Swarm Optimization}
PSO belongs to the class of swarm intelligence techniques and is a population-based stochastic technique for solving optimization problems developed in 1995.  An open source python library, for CNN optimization using the PSO algorithm was developed by Fernandes et al \cite{junior2019particle}. The results demonstrate that their approach, psoCNN, quickly finds CNN architectures which offer competitive performance for any given dataset. 

\subsubsection{Ant Colony Optimization}

ACO, modelled on the activities of real ant colonies, involves moving through a parameter space of all potential solutions to find the optimal weights for a neural network. 
\par
Using ACO, a system known as DeepSwarm was developed by Byla and Pang \cite{byla2019deepswarm} to find high performing neural architectures for CNNs. They showed that it offers competitive performance when tested on well-known image datasets.

\section{Approach}

\par
With artificial neural networks, there are many parameters to choose from such as the number of hidden network layers, number of neurons per layer,  type of activation function, choice of optimizer and so on. The final network design often depends on the problem domain and is typically achieved in a time consuming trial and error fashion.
\par
Similar problems exist with CNNs but these problems are exacerbated by the length of time, and amount of computational resources required to train such networks. Clearly, a core objective of NAS is to find good network performance within acceptable time limits through the reduction of both the number of networks tested and the length of time required for their evaluation. The implementation of NAS can be achieved through a variety of approaches including transfer learning using pre-trained networks, network morphism or swarm intelligence. Using these approaches as its pillars, a NAS system (OpenNAS) has been built which tackles such problems \footnote{ https://github.com/seamusl/OpenNAS-v1}. OpenNAS does not enforce a particular architecture but rather it allows novel and interesting architectures to be discovered. 
\par
In this work we focus on the swarm intelligence component of the OpenNAS system. The swarm optimization techniques currently used are Particle Swarm Optimization and Ant Colony Optimization. 
\begin{table}[htb]
\centering
\caption{Parameters for Particle Swarm Optimization}
\begin{tabular}{|c|c|l|}
\hline
\textbf{}                                    & \textbf{Config A}  & \textbf{Config B}  \\ \hline
\multicolumn{3}{|c|}{\cellcolor[HTML]{C0C0C0}\textit{Swarm}}     \\ \hline
Number of iterations                         & 10                 & 20                 \\ \hline
Swarm size                                   & 20                 & 10                 \\ \hline
Cg                                           & 0.5                & 0.5                \\ \hline
\multicolumn{3}{|c|}{\cellcolor[HTML]{C0C0C0}\textit{CNN architecture}} \\ \hline
Minimum outputs from a Conv layer  & 3                  & 3                  \\ \hline
Maximum outputs from a Conv layer  & 256                & 256                \\ \hline
Maximum neurons in a FC layer      & 300                & 300                \\ \hline
Minimum size of a Conv kernel                & 3 x 3              & 3 x 3              \\ \hline
Maximum size of a Conv kernel                & 7 x 7              & 7 x 7              \\ \hline
Minimum layers                     & 3                  & 3                  \\ \hline
Maximum layers                     & 20                 & 20                 \\ \hline
\multicolumn{3}{|c|}{\cellcolor[HTML]{C0C0C0}\textit{CNN Training}}                    \\ \hline
\# epochs for particle evaluation            & 5                  & 5                  \\ \hline
\# epochs for global best                & 100                & 100                \\ \hline
Dropout rate                                 & 0.5                & 0.5                \\ \hline
Batch normalize layer outputs                & Yes                & Yes                \\ \hline
\multicolumn{3}{|c|}{\cellcolor[HTML]{C0C0C0}\textit{Probability Settings}}            \\ \hline
probability\_convolution                     & 0.6                & 0.6                \\ \hline
probability\_pooling                         & 0.3                & 0.3                \\ \hline
probability\_fully\_connected                & 0.1                & 0.1                \\ \hline
\end{tabular}
\label{table:psoParameters}
\end{table}

\par
The PSO algorithm determines how the principal CNN layer types, and their associated hyperparameters, are connected together. The generated models consist of architectures using a mix of convolutional, average pooling, max pooling and fully connected layers.  In addition, dropout layers and batch normalization layers are also added to alleviate overfitting.  The hyperparameters associated with each layer type are indicated in Table ~\ref{table:psoParameters}.  
\par
Particle architectures, i.e. model architectures, are compiled for a number of epochs and evaluation is carried out using the standard loss function of cross-entropy loss. Particle architectures with the smallest loss are selected by the algorithm.  The number of epochs parameter for pBest must be carefully chosen since it is the main driver of both run time and model accuracy.  
\par
Using an ACO approach, the parameters used for model training in the exploration process are highlighted in Table 2. Two test configurations are considered. In the first case, 8 ants are used with 30 epochs and in the second case, 16 ants are used with 15 epochs. The depth parameter was fixed at 20. 
\begin{table}[htb]
\label{table:acotable}
\centering
\caption{Parameters for Ant Colony Optimization}
\begin{tabular}{|c|c|l|}
\hline
\textbf{}                                    & \textbf{Config A}  & \textbf{Config B}  \\ \hline
\multicolumn{3}{|c|}{\cellcolor[HTML]{C0C0C0}\textit{Ant Colony}}     \\ \hline
Number of Ants                         &8                 & 16                 \\ \hline
Number of Epochs                                   & 30                 & 15                 \\ \hline
Search Depth                                           & 20                & 20                \\ \hline
\multicolumn{3}{|c|}{\cellcolor[HTML]{C0C0C0}\textit{CNN architecture}} \\ \hline
Kernel Sizes               & 1, 3, 5              & 1, 3, 5              \\ \hline
Minimum layers                     & 1                  & 1                  \\ \hline
Maximum layers                     & 20                 & 20                 \\ \hline
\multicolumn{3}{|c|}{\cellcolor[HTML]{C0C0C0}\textit{CNN Training}}                    \\ \hline
Dropout rate                                 & 0.1, 0.3, 0.5                & 0.1, 0.3, 0.5                \\ \hline
Batch normalize layer outputs                & Yes                & Yes                \\ \hline
\multicolumn{3}{|c|}{\cellcolor[HTML]{C0C0C0}\textit{Probability Settings}}            \\ \hline
pheromone start, decay, evaporation          & 0.1                & 0.1                \\ \hline
greediness                                   & 0.5                & 0.5                \\ \hline
\end{tabular}
\end{table}
\par
\begin{figure}
\centering
\includegraphics[scale=0.2]{{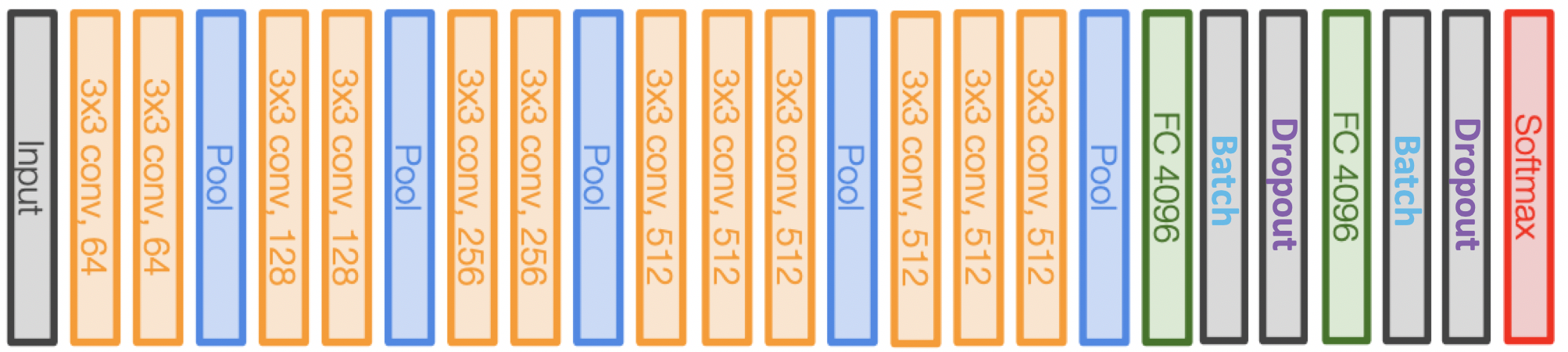}}
  \caption{Fine-tuned VGG16 model}
  \label{fig:tuned}
\end{figure}

Fine-tuning was implemented by initially removing the fully connected layers from the top of the model. Two blocks are then added, each of which has a fully connected layer, a batch normalization layer and a dropout layer. The hybrid structure is then trained with the new dataset. Fine-tuning of a VGG16  network is illustrated in Figure~\ref{fig:tuned}. 

\section{Design}

\begin{figure}[ht]
  \noindent
  \makebox{\includegraphics[width=\columnwidth]{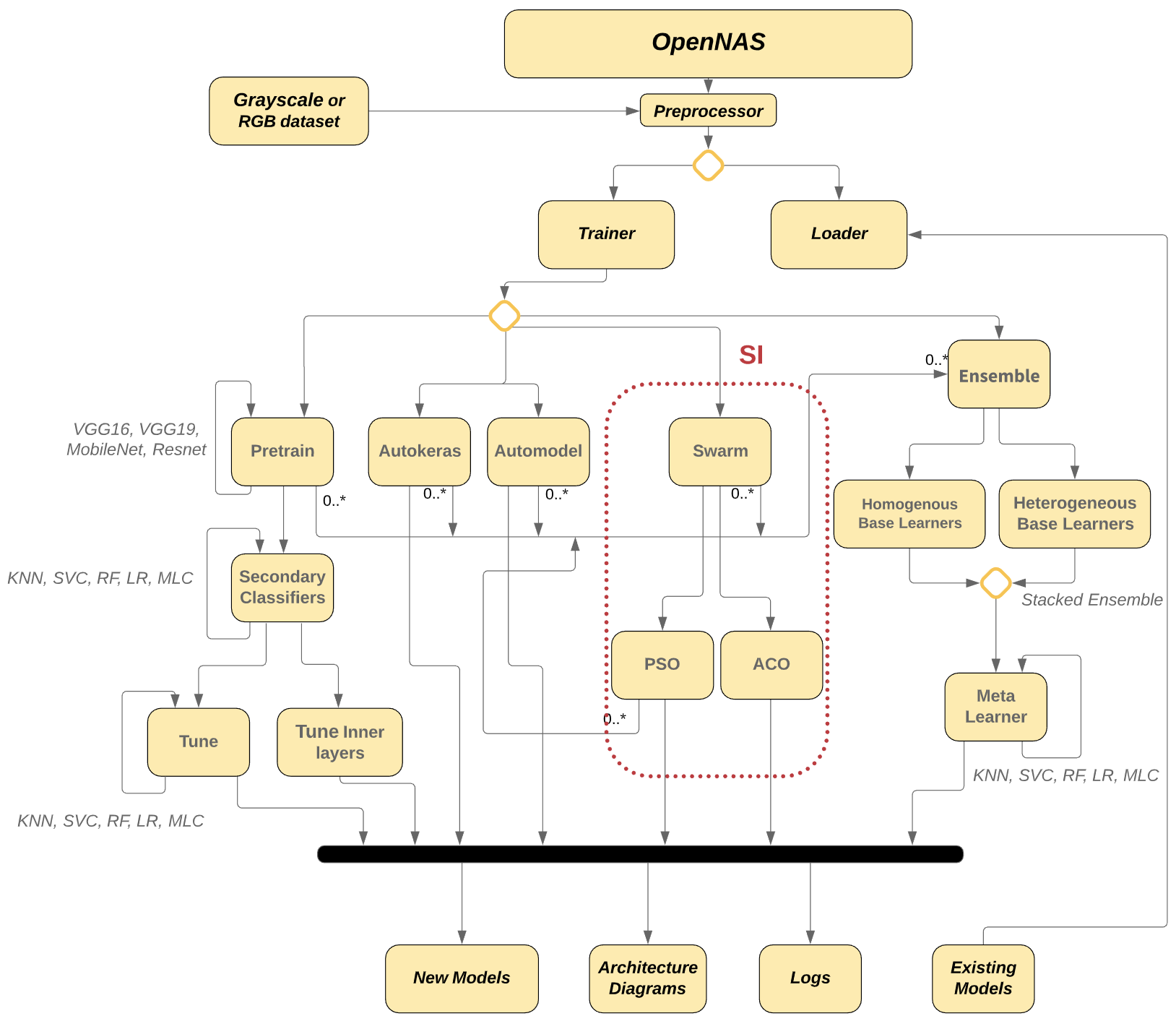}}%
  \caption{Open source Neural Architecture Search (OpenNAS) system design}
  \label{fig:OpenNAS}
\end{figure}

The high level view of the system architecture is presented in Figure~\ref{fig:OpenNAS}. The system is organized into the following python modules: OpenNAS, pre-processor, trainer, ensemble, super stacker, sysconfig and loader. The pre-train function uses transfer learning as either a feature extractor or to fine-tune the pre-trained networks of VGG16, VGG19, MobileNet or ResNet50. 
\par
With the swarm function, PSO or ACO can be used to search for the best neural architecture. Existing open source python libraries were customized for both PSO and ACO functionality. Particle swarms were implemented using a psoCNN library\cite{junior2019particle} whereas ant colonies used the DeepSwarm library \cite{byla2019deepswarm}. The environment required Python 3.7, Tensorflow 1.14,  Keras 2.2.4, Numpy 1.16.4 and Matplotplib 3.1.0.

\par
Existing NAS tools, such as AutoKeras, were also integrated into the OpenNAS system. AutoKeras is a powerful open source library which provides functions to automatically search for optimal architectures for deep learning models.  However, this library is still in beta development and the associated documentation is quite poor.   
\par
With the ensemble module, there are options to build stacked ensembles using either homogeneous or heterogeneous base learners. These learner outputs are subsequently passed to a suite of meta learner algorithms. The system generates the optimal neural architecture model using the chosen heuristic.

\section{Evaluation}
\par
Two datasets were chosen for the experimental design, namely CIFAR10 \cite{krizhevsky2014cifar} and Fashion\_Mnist \cite{xiao2017fashion}. A primary research objective is the development of a neural architecture search tool which generates high performing architectures for generic datasets of either grayscale (one channel) or color (triple channel) images. The CIFAR10 dataset meets this requirement in that it is a challenging dataset of color images. The Fashion\_Mnist dataset is also suitable since it a well-tested and well-understood dataset of black and white images. For reference, the state of the art (SOA) accuracy achieved on CIFAR10 is 98.5\%\cite{cubuk2019autoaugment} whereas with Fashion\_Mnist, the SOA accuracy is 94.6\%\cite{ma2020autonomous}.
  
\subsection{Particle Swarm Optimization }
In order to test variance and reproducibility, each configuration was run 5 times on both CIFAR10 and Fashion\_Minst which resulted in the evaluation of 4000 CNN architectures for this phase of the study. 

\subsubsection{Evaluation of models trained on CIFAR10 dataset}

Validation accuracy was used to evaluate the performance of both PSO configurations. It is clear from Table~\ref{tab:pso-cifar10} that the PSO model trained on swarm settings of a lower population and higher number of iterations (population of 10 and 20 iterations) performed significantly better. In terms of accuracy, the mean performance was 3.5\% better.   

\begin{table}[]
\centering
\caption{Performance of PSO models on CIFAR10}
\label{tab:pso-cifar10}
\begin{tabular}{lccccc}
\hline
\multicolumn{1}{c}{\textbf{Model}} &
  \textbf{\begin{tabular}[c]{@{}c@{}}Acc \\ Max \hspace{0.5mm}\end{tabular}} &
  \textbf{\begin{tabular}[c]{@{}c@{}}Acc\\ Mean \hspace{0.5mm}\end{tabular}} &
  \textbf{\begin{tabular}[c]{@{}c@{}}Acc\\ StDev \hspace{0.5mm}\end{tabular}} &
  \textbf{\begin{tabular}[c]{@{}c@{}}Time \\ (min)\hspace{0.5mm}\end{tabular}} &
  \textbf{Layers} \\ \hline
\textbf{\begin{tabular}[c]{@{}l@{}}Population 10 \\ Iterations \hspace{0.5mm} 20 \end{tabular}} &
  0.900 &
  0.853 &
  0.044 &
  1316 &
  30 \\ \hline
\textbf{\begin{tabular}[c]{@{}l@{}}Population 20 \\ Iterations \hspace{0.5mm} 10 \end{tabular}} &
  0.883 &
  0.818 &
  0.053 &
  1119 &
  38 \\ \hline
\end{tabular}
\end{table}

\par
Both configurations have a very low standard deviation for model accuracy indicating a high level of reproducibility between test runs. At a mean run time of 21.9 hours for the first configuration and 18.6 hours for the second configuration, the PSO search for CNN architectures is a slow process considering high performance workstations, with NVIDIA GeForce GTX 1080 Ti graphic cards, were used. 

\subsubsection{Evaluation of models trained on Fashion\_Mnist dataset}
\par
PSO models trained on the Fashion\_Mnist dataset (Table~\ref{tab:pso-fashion}), achieved much higher accuracy compared with models developed using CIFAR10 data. Similar to CIFAR10, the low standard deviation associated with both implementations of Fashion\_Mnist models indicate the PSO approach produces consistent results between different test runs.

\begin{table}[]
\caption{Performance of PSO models on Fashion\_Mnist}
\centering
\label{tab:pso-fashion}
\begin{tabular}{lccccc}
\hline
\multicolumn{1}{c}{\textbf{Model}} &
  \textbf{\begin{tabular}[c]{@{}c@{}}Acc\\ Max\hspace{0.5mm}\end{tabular}} &
  \textbf{\begin{tabular}[c]{@{}c@{}}Acc\\ Mean\hspace{0.5mm}\end{tabular}} &
  \textbf{\begin{tabular}[c]{@{}c@{}}Acc\\ StDev\hspace{0.5mm}\end{tabular}} &
  \textbf{\begin{tabular}[c]{@{}c@{}}Time \\ (min)\hspace{0.5mm}\end{tabular}} &
  \textbf{Layers} \\ \hline
\textbf{\begin{tabular}[c]{@{}l@{}}Population: 10 \\ Iterations: \hspace{0.5mm} 20\end{tabular}} &
  0.943 &
  0.932 &
  0.009 &
  994 &
  30 \\ \hline
\textbf{\begin{tabular}[c]{@{}l@{}}Population: 20 \\ Iterations: \hspace{0.5mm} 10\end{tabular}} &
  0.943 &
  0.935 &
  0.008 &
  1319 &
  38 \\ \hline
\end{tabular}
\end{table}
\par
The stochastic nature of metaheuristics impacts the run times associated with PSO for both Fashion\_Mnist and CIFAR10. In all tests, no clear pattern emerged with regard to run times: CIFAR10 was faster using a population of 10 with 20 iterations whereas Fashion\_Mnist was faster with a population of 20 with 10 iterations. Therefore, in terms of run time, no clear conclusion could be drawn by doubling the population and halving the iterations.
\par
With regard to the impact of swarm settings on model accuracy for Fashion\_Mnist, again there is little to separate the configurations. With a mean accuracy of 93.5\% for a population of 20 with 10 iterations and a corresponding mean model accuracy of 93.2\% using a population of 10 with 20 iterations, no clear conclusion can be drawn.  
\par
Therefore, unlike CIFAR10, changing the swarm settings by doubling population and halving iterations does not impact model accuracy in the case of Fasion\_Mnist. Both configurations for the PSO algorithm perform well on this dataset.  

\subsection{Ant Colony Optimization }

\par
Similar to other metaheuristics, there are several parameters which can be tuned for optimal neural architecture search using Ant Colony Optimization \cite{dorigo1997ant}. With OpenNAS, users may select the options of depth, number of ants and number of epochs in directing how the neural architecture search is conducted. 

\subsubsection{Evaluation of models trained on CIFAR10 dataset}
With CIFAR10 data, the results from Table~\ref{tab:aco-cifar} indicate that a greater number of ants leads to higher model accuracy. The improvement in max model accuracy achieved, through doubling the number of ants and halving the number of epochs, was modest at just 1.2\%. The impact on run time for a small increase in accuracy was severe. Doubling the number of ants effectively doubled the run time (even though the number of epochs was halved). The standard deviation for accuracy is very low indicating good reproducibility between the various test runs.
\begin{table}[]
\centering
\caption{Performance of ACO models on CIFAR10}
\label{tab:aco-cifar}
\begin{tabular}{lccccc}
\hline
\multicolumn{1}{c}{\textbf{Model}} &
  \textbf{\begin{tabular}[c]{@{}c@{}}Acc\\ Max \hspace{0.5mm}\end{tabular}} &
  \textbf{\begin{tabular}[c]{@{}c@{}}Acc\\ Mean\hspace{0.5mm}\end{tabular}} &
  \textbf{\begin{tabular}[c]{@{}c@{}}Acc\\ StDev\hspace{0.5mm}\end{tabular}} &
  \textbf{\begin{tabular}[c]{@{}c@{}}Time \\ (min)\hspace{0.5mm}\end{tabular}} &
  \textbf{Layers} \\ \hline
\textbf{\begin{tabular}[c]{@{}l@{}}Ants: \hspace{5mm}8 \\Epochs: 30\end{tabular}} &
  0.848 &
  0.822 &
  0.025 &
  541 &
  18 \\ \hline
\textbf{\begin{tabular}[c]{@{}l@{}}Ants: \hspace{4mm}16\\Epochs: 15\end{tabular}} &
  0.836 &
  0.821 &
  0.014 &
  1004 &
  16 \\ \hline
\end{tabular}
\end{table}

\subsubsection{Evaluation of models trained on Fashion\_Mnist dataset}

\par
The performance of ACO models using Fashion\_Mnist data is highlighted Table~\ref{tab:aco-fashion}. It can be seen that both configurations perform well resulting in accuracies greater than 93\%. The difference in mean model accuracy between configuration A (8 ants and 30 epochs) is trivial when compared to configuration B (16 ants and 15 epochs). However, similar to ACO on CIFAR10, the difference in run time is very significant for configuration B.  Effectively it took over 7 hours longer to achieve an accuracy improvement of 0.1\%. 
\par
Clearly in the case of a simpler dataset such as Fashion\_Mnist, using a number of ants in excess of 8 is not worth doing. This finding is similar to that seen with the more complex CIFAR10 dataset, above. Therefore choosing the number of ants, used for this ACO implementation, is an important consideration impacting run time performance.  As anticipated, the standard deviation for accuracy is also very low indicating good reproducibility between the various test runs.

\begin{table}[]
\centering
\caption{Performance of ACO models on Fashion\_Mnist}
\label{tab:aco-fashion}
\begin{tabular}{lccccc}
\hline
\multicolumn{1}{c}{\textbf{Model}} &
  \textbf{\begin{tabular}[c]{@{}c@{}}Acc\\ Max \hspace{0.5mm}\end{tabular}} &
  \textbf{\begin{tabular}[c]{@{}c@{}}Acc\\ Mean\hspace{0.5mm}\end{tabular}} &
  \textbf{\begin{tabular}[c]{@{}c@{}}Acc\\ StDev\hspace{0.5mm}\end{tabular}} &
  \textbf{\begin{tabular}[c]{@{}c@{}}Time \\ (min)\hspace{0.5mm}\end{tabular}} &
  \textbf{Layers} \\ \hline
\textbf{\begin{tabular}[c]{@{}l@{}}Ants: \hspace{5mm}8 \\Epochs: 30\end{tabular}} &
  0.934 &
  0.931 &
  0.002 &
  375 &
  7 \\ \hline
\textbf{\begin{tabular}[c]{@{}l@{}}Ants: \hspace{4mm}16 \\Epochs: 15\end{tabular}} &
  0.934 &
  0.932 &
  0.004 &
  837 &
  19 \\ \hline
\end{tabular}
\end{table}

\section{Discussion}

\begin{figure}[htb]
  \centering
\includegraphics[width=\columnwidth]{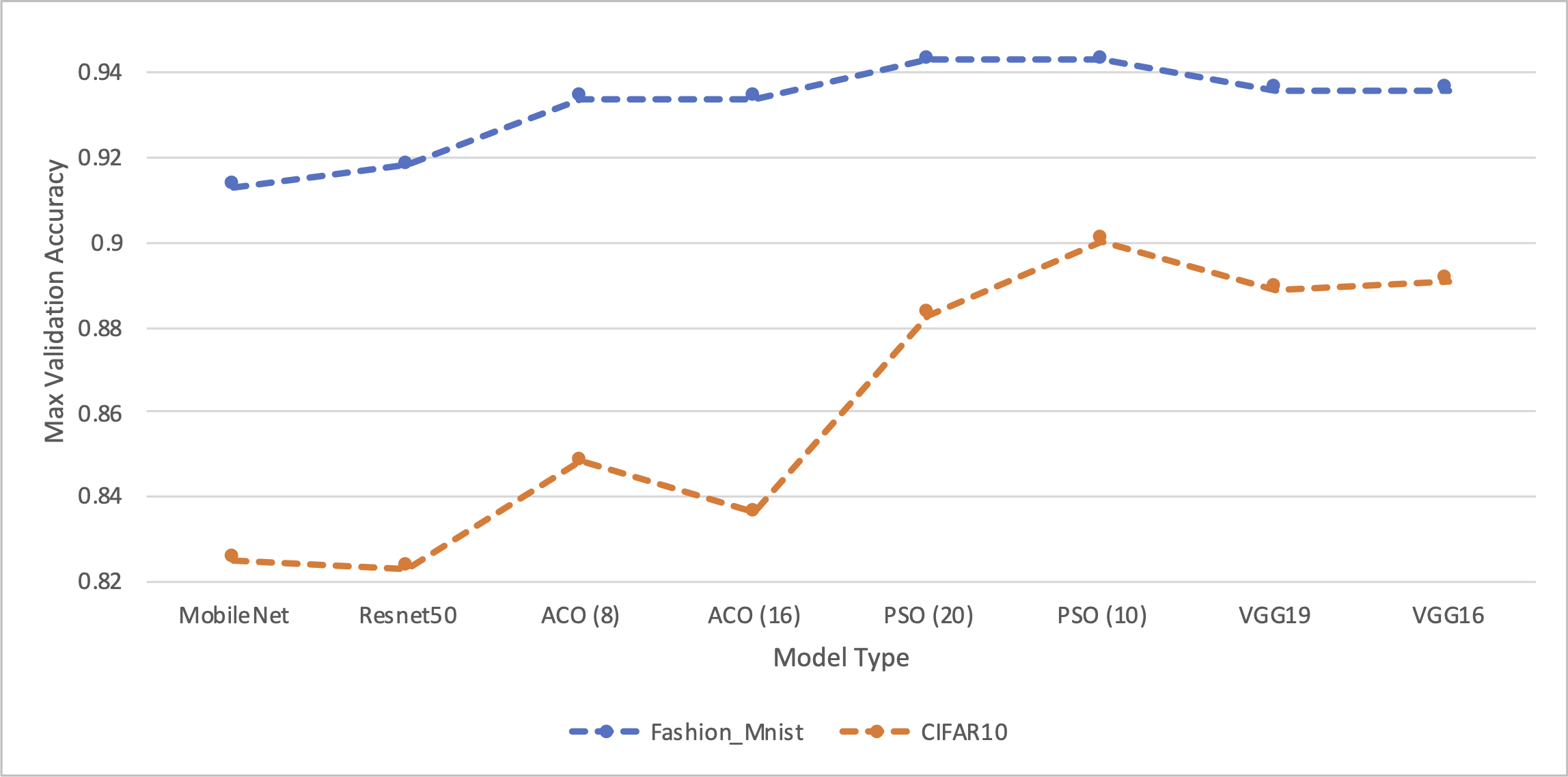}
  \caption{OpenNAS performance of all models on CIFAR10 and Fashion\_Mnist}
  \label{fig:compare-all}
\end{figure}

\par
The OpenNAS performance of all models across both datasets is illustrated in Figure~\ref{fig:compare-all}. The results demonstrate performance comparable to that achieved by the psoCNN \cite{junior2019particle} approach and better than that of DeepSwarm \cite{byla2019deepswarm}. 
\par
The highest accuracy of OpenNAS in CIFAR-10 classification was 90.0\%. This was achieved using using a PSO-derived model. By comparison, DeepSwarm achieved a top accuracy of 88.7\%.
\par
With Fashion\_Mnist data, the highest performing model for OpenNAS is again a PSO derived model with an accuracy of 94.3\%. This result compares very favourably with the SOA accuracy of 94.6\%. The highest performing model for DeepSwarm achieved an accuracy of 93.56\%. 
\par
In the case of psoCNN, experiments were conducted on Fashion\_Mnist but not on CIFAR-10. The best performing psoCNN model on Fashion\_Mnist was 91.9\% without dropout and 94.5\% with dropout.    
\par
The findings clearly show that a PSO approach leads to higher model accuracies given that DeepSwarm is exclusively based on an ACO approach.
\par
The pre-trained networks of MobileNet and RestNet50 delivered the poorest performance with CIFAR10. The other pre-trained networks, using VGG architectures, performed very well on the same dataset. 
\par
With a more complex dataset, such as CIFAR10, the mean performance improvement of the PSO algorithm is significant when compared with ACO. With configurations used in this study, PSO achieved a mean accuracy of 85.3\% on CIFAR10 compared with an ACO mean accuracy of 82.2\%. 
\par
The approach taken by ACO, in determining the best architecture is very different to the PSO approach. With ACO, simpler models are initially evaluated at lower depths with progressively more complex models being evaluated at deeper search levels. Therefore at search depth 1, there is essentially just a single hidden layer being evaluated. The number of ants specified creates new architectures which simply vary the hyper parameters used for that layer.  With each new depth being explored, an additional layer is added to the architecture being explored.
\par
Furthermore, the ACO approach enables the targeting of hyper parameter optimization within a given layer type rather than optimizing at the overall architecture level. Specifying a large number of ants, with a reduced depth, ensures the search space is restricting to studying the effects of layer hyper parameters rather than model depth and the constituent layers. By comparison, the number of layers in PSO generated models is entirely stochastic. 

\section{Conclusion}
\par
The OpenNAS approach identifies the hyperparameters within each layer of networks used for image classification of grayscale and color datasets. In addition the number and type of layers for the neural architecture are also identified. This combined approach generates model architectures which achieve competitve accuracies when classifying the CIFAR10 and Fashion\_Mnist datasets.
\par
The results of swarm intelligence algorithms, in the context of this study, have generated impressive performances. However, in many cases, their performance is only marginally better than fine-tuned pre-trained VGG models. 
The accuracies of PSO derived models have been shown to exceed those of ACO derived models in the image classification of grayscale and color datasets. 
\par
In addition, the OpenNAS integrated approach, using both PSO and ACO algorithms, yields higher accuracies when compared with DeepSwarm which relies on a single metaheuristic. 

\bibliographystyle{IEEEtran}
\bibliography{mybib}

%
%
%

\end{document}